\documentclass[10pt,a4paper,conference]{IEEEtran}

\usepackage{graphicx}
\graphicspath{{images/}}
\DeclareGraphicsExtensions{.pdf,.jpeg,.png}

\usepackage{amsmath}
\usepackage{amssymb}
\usepackage{multirow}
\usepackage{booktabs}
\usepackage{array}
\usepackage{color}
\usepackage[flushleft]{threeparttable}

\usepackage{cite}

\ifCLASSOPTIONcompsoc
 \usepackage[caption=false,font=normalsize,labelfont=sf,textfont=sf]{subfig}
\else
 \usepackage[caption=false,font=footnotesize]{subfig}
\fi

\usepackage{url}

\begin{document}
%
\title{DeepFirearm: Learning Discriminative Feature Representation for Fine-grained Firearm Retrieval}



%
\author{\IEEEauthorblockN{Jiedong Hao\IEEEauthorrefmark{1}\IEEEauthorrefmark{2},
Jing Dong\IEEEauthorrefmark{1},
Wei Wang\IEEEauthorrefmark{1} and
Tieniu Tan\IEEEauthorrefmark{1}}

\IEEEauthorblockA{\IEEEauthorrefmark{1} Center for Research on Intelligent Perception and Computing (CRIPAC),\\Institute of Automation, Chinese Academy of Sciences}
\IEEEauthorblockA{\IEEEauthorrefmark{2}  University of Chinese Academy of Sciences}
\texttt{jiedong.hao@cripac.ia.ac.cn, \{wwang, jdong, tnt\}@nlpr.ia.ac.cn}
}


\maketitle

\begin{abstract}
There are great demands for automatically regulating inappropriate appearance of shocking firearm images in social media or identifying firearm types in forensics. Image retrieval techniques have great potential to solve these problems. To facilitate research in this area, we introduce Firearm 14k, a large dataset consisting of over 14,000 images in 167 categories. It can be used for both fine-grained recognition and retrieval of firearm images. Recent advances in image retrieval are mainly driven by fine-tuning state-of-the-art convolutional neural networks for retrieval task. The conventional single margin contrastive loss, known for its simplicity and good performance, has been widely used. We find that it performs poorly on the Firearm 14k dataset due to: (1) Loss contributed by positive and negative image pairs is unbalanced during training process. (2) A huge domain gap exists between this dataset and ImageNet. We propose to deal with the unbalanced loss by employing a double margin contrastive loss. We tackle the domain gap issue with a two-stage training strategy, where we first fine-tune the network for classification, and then fine-tune it for retrieval. Experimental results show that our approach outperforms the conventional single margin approach by a large margin (up to 88.5\% relative improvement) and even surpasses the strong triplet-loss-based approach.
\end{abstract}

%
\IEEEpeerreviewmaketitle

\section{Introduction}
In social media such as Facebook, there are huge number of posts with firearm images~\cite{Forbes2016FacebookGun}, which may need proper regulation. In forensic science, sometimes a forensic examiner may need to identify the types and models of firearms from a large dataset of firearm images. Image retrieval methods can be helpful in both cases. In this paper, we focus on the problem of fine-grained retrieval of firearm images, i.e., given a query image, find the database images belonging to the same finer category. This task is difficult as both the training and testing images are ``in the wild'', i.e., the images are not aligned and object instances may appear in various scales and in different positions of the images.

Few study has been conducted on the topic of firearm retrieval. A large dataset with various firearm types is needed for further research in this area. This work aims to fill the gap. To this aim, we have collected \emph{Firearm 14k}, a high quality firearm dataset, which contains 14,755 images in 167 categories. This is a challenging dataset for firearm retrieval due to large variations in object pose, orientation, occlusion, background clutter etc. (see Fig.~\ref{fig:dataset_imgs} for some example images in this dataset). We will make it publicly available to the research community\footnote{Dataset is available at \url{http://forensics.idealtest.org/Firearm14k/}}. We believe that it will facilitate future researches on firearm recognition and retrieval.

With the great success of convolutional neural networks (CNNs)~\cite{Krizhevsky2012ImageNetCW} on ImageNet~\cite{Russakovsky2015ImageNetLS} 1k-class classification task, it has been shown that fine-tuning the CNNs end-to-end for other tasks such as object detection~\cite{Ren2015FasterRT} and semantic segmentation~\cite{Shelhamer2015FullyCN} can achieve remarkable results. Following this trend, researchers in the image retrieval community have also adopted this technique. Radenovic et al.~\cite{Radenovic2016CNNIR} fine-tuned a two stream Siamese network with a contrastive loss~\cite{Chopra2005LearningAS}, while Gordo et al.~\cite{Gordo2016DeepIR} utilized a three stream network with triplet loss~\cite{Wang2014LearningFI,Schroff2015FaceNetAU} for feature learning. Both their aims are to learn feature embeddings tailored for image retrieval task through end-to-end training. Their results demonstrate that end-to-end learning significantly improves the performance of the model on standard datasets.

The Siamese network~\cite{Radenovic2016CNNIR,Chopra2005LearningAS,Bell2015LearningVS} has been very popular in metric-learning-based approaches. It tries to learn an embedding space where feature distances between similar images are small, while feature distances between dis-similar images are large. However, a noticeable drawback of this approach is that the contrastive loss used is unbalanced: the network will penalize positive image pairs even if their feature distances are small enough. The force to pull the positive image pairs together is so strong that performance of the model is compromised. Another problem with existing methods is that the retrieval datasets~\cite{Philbin2007ObjectRW,Philbin2008LostIQ} are often similar in style to the ImageNet dataset, where the classification models are trained. As a result, retrieval accuracy on these datasets are relatively high even if off-the-shelf models are utilized~\cite{Tolias2015ParticularOR,Babenko_2015_ICCV,Babenko2014NeuralCF}. For datasets which are vastly different from ImageNet (e.g., the Firearm 14k dataset), even the fine-tuned models may work poorly.

In this work, we propose to use a double margin contrastive loss to train the CNNs end-to-end, in order to solve the problem of unbalanced loss. In this loss, we set separate margins for the positive and negative image pairs, thus striking a better balance between the positive and negative pairs during training process. The margin terms are selected based on the feature distance distributions of positive and negative image pairs. Experimental results in Sec.~\ref{subsec:single_vs_double_exp} show that our double margin approach outperforms the conventional single margin approach by a large margin (34.2\%) with the VGG16 base model~\cite{Simonyan2014VeryDC}, with no bells and whistles. A two-stage training strategy is developed to deal with the large domain gap between Firearm 14k and ImageNet. Specifically, we first fine-tune the model for classification on the Firearm 14k dataset, and then fine-tune the model for retrieval with the double margin loss. This strategy proves beneficial for further boosting the performance of our model.

We make three \textbf{contributions} in this work. (1) We build Firearm 14k, a large dataset consisting of over 14,000 firearm images in 167 categories, which is the largest of its kind. (2) We propose to learn discriminative global feature embeddings for retrieval with a double margin contrastive loss, which deals with the unbalanced loss arising from the use of single margin method. (3) We use a two-stage training strategy to alleviate the domain gap between Firearm 14k and ImageNet. Overall, our model shows superior performance when compared to the state-of-the-art approaches on firearm retrieval task.

\section{Related Work}
Wen and Yao~\cite{Wen2005PistolIR} propose to retrieve similar pistol images using their edges to centers distance distributions. The difference of distributions is used to measure the similarity of two pistols. Their pistol dataset has a small number of around 300 images.

Early work~\cite{Razavian2014CNNFO,Babenko2014NeuralCF} employing CNNs for retrieval treat the output of fully-connected layers as image features. Later, Babenko and Lempitsky~\cite{Babenko_2015_ICCV} perform sum pooling over the output feature maps of last convolutional layer to generate a global image representation, while Tolias et al.~\cite{Tolias2015ParticularOR} propose to use max pooling aggregation instead (denoted as \emph{MAC}) and obtain better results. All these work does not perform end-to-end learning for the specific dataset and treat the CNNs as off-the-shelf feature extractors. These methods will not perform well when the target dataset is significantly different from ImageNet on which CNNs are trained. All these work except Tolias et al.~\cite{Tolias2015ParticularOR} requires that the image size should be fixed, which will also harm the retrieval performance as the images may lose some detail information when they are warped~\cite{kaiming14ECCV,Hao2017MFCAM}. In our work, a fully-convolutional network with the MAC feature representation~\cite{Tolias2015ParticularOR} is utilized, and we train the network end-to-end to optimize it for our specific dataset.

Babenko et al.~\cite{Babenko2014NeuralCF} fine-tuned the AlexNet~\cite{Krizhevsky2012ImageNetCW} using a collected landmark dataset with classification loss and observe improved results on retrieval task. But the cross entropy loss used is not directly targeted at retrieval task and cannot generalize well to unseen classes. Gordo et al.~\cite{Gordo2016DeepIR} thus propose to directly learn image feature embeddings targeted at retrieval task with a triplet loss. Concurrent to the work of Gordo et al., Radenovic et al.~\cite{Radenovic2016CNNIR} propose to directly optimize the CNN networks with the single margin contrastive loss. They find that the Siamese network works better than the triplet network. In our work, we also use a two branch Siamese network. However, we find that the single margin contrastive loss is less effective for Firearm 14k even with fine-tuning. We propose to deal with this issue using a double margin contrastive loss instead. Cao et al.~\cite{Cao2016QuartetnetLF} also employ a similar form of the double margin loss for the problem of visual instance retrieval.

\begin{table}[t]
  \centering
  \caption{The number of classes where each labeler has the lowest average correlation.}
  \label{table:largest_devi_num}
  \begin{tabular}{@{}llllll@{}}
    \toprule
    Labeler ID  &  0    & 1 & 2 & 3 & 4  \\
    \midrule
    Number &  4 & 93 & 4 & 33 & 33 \\
    \bottomrule
  \end{tabular}
\end{table}

\section{The Firearm 14k Dataset}
In this section, we briefly introduce the image collection, labeling and filtering process of the Firearm 14k dataset. Finally, we give a summary of the dataset statistics.

\subsection{Image Collection and Cleaning}
First, we compile a list of keywords from websites and forums, which correspond to different firearms types such as \emph{AK-47} and \emph{Glock 17}. In total, we manage to gather 167 keywords. We then use these keywords to search relevant images through the Google image search engine. For each keyword, 300--500 images are collected. Upon finishing image collection, we remove either gray-scale, small or non-JPEG images.

\subsection{Image Labeling and Filtering}
\noindent\textbf{Image Labeling.}
To filter out noisy images, we employ five labelers to give similarity scores (in the range [0, 9], similarity increases from 0 to 9) to images in each class based on their similarities to the given exemplar images. Each labeler independently scores all the dataset images once.

\noindent\textbf{Filtering Valid Images.} It is inevitable that the labeling quality of the five labelers is uneven. In order to assign a reliable score to each image, we need to assess the labelers' labeling quality properly. The labeling score correlation is used. Ideally, for each class, the score correlation between two labelers should be high, although the exact score for an image may vary. If a labeler gives low-quality scores, the correlation between his score and others' will be low. We calculate the number of classes where each labeler has the lowest average correlation. The smaller that number, the better a labler's labeling quality. We show the statistics in Table~\ref{table:largest_devi_num}. It is noticeable that labeler 1 has poor labeling quality.

For each image, we use a weighted average of the five scores to compute its final score, which is computed as $\sum_{i=0}^{4}w_{i}s_{i}$, where $s_i$ denotes the labeler $i$'s score and $w_i$ is the corresponding weight. The weight we use is $w = [0.445, 0, 0.445, 0.055, 0.055]$. We set the weight of labeler 1 to 0 becaue of the poor labeling quality. For each class, we select images whose final scores satisfy $s \geq 6$ as valid images and collect all such images to form our final dataset.

\subsection{Dataset Statistics}\label{subsec:dataset_stat}
\begin{figure}[t]
  \centering
  \includegraphics[width=\linewidth]{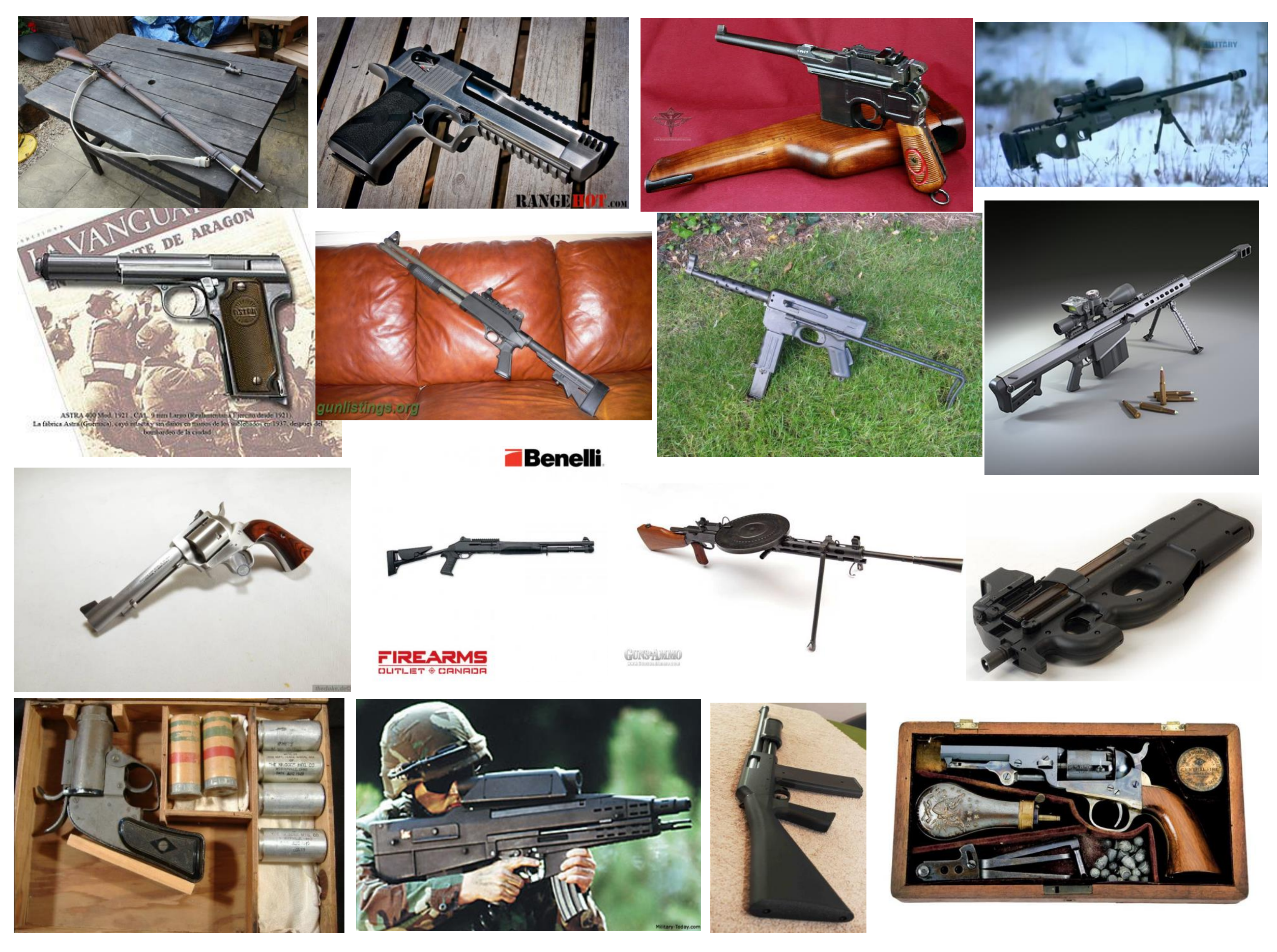}
  \caption{Example images from the Firearm 14k dataset.}
  \label{fig:dataset_imgs}
\end{figure}

\begin{table}[t]
  \centering
  \caption{Statistics of the Firearm 14k dataset.}
  \label{table:train_val_test_stat}
  \begin{tabular}{@{}lllll@{}}
    \toprule
              & \# classes & \# images & \# queries  &\# database images \\
    \midrule
    Train set & 107       & 9,628          & --  & --     \\
    Validation set & 20    & 1,478           & 39 & 1,438     \\
     Test set & 40       & 3,649          & 80    & 3,569   \\
     Total &  167 &  14,755  & -- & -- \\
    \bottomrule
  \end{tabular}
\end{table}

The final Firearm dataset consists of 14,755 images from 167 classes of various firearm types. It has two characteristics worth noting: (1) This dataset is highly unbalanced: the number of images in a class varies from 19 to over 200. (2) Images from the Firearm 14k dataset have large variations in object scale, orientation, pose, lighting etc., which makes the dataset challenging for recognition. We show some example images from the dataset in Fig.~\ref{fig:dataset_imgs}.

We split the dataset into train, validation and test set, each containing about 65\%, 10\% and 25\% of the whole images. To mimic the real world retrieval scenario, the three sets have disjoint classes. For the validation and test set, we randomly pick two images from each class as query images\footnote{One query image from the validation set is removed because of mislabeling.} and use the remaining images as database images.
We summarize the detailed statistics of the Firearm 14k dataset in Table~\ref{table:train_val_test_stat}.

\subsection{Performance Evaluation}
We evaluate the retrieval performance under two metrics on the test set. The first one is \emph{mean average precision }(mAP)~\cite{Philbin2007ObjectRW}, which is used to measure overall performance of retrieval systems. The second one is \emph{rank-k accuracy}, which calculates the percentage of query images whose $k$-nearest neighbors contain at least one image from the same class.

\section{Our Approach}
\begin{figure}[t]
  \centering
  \includegraphics[width=\linewidth]{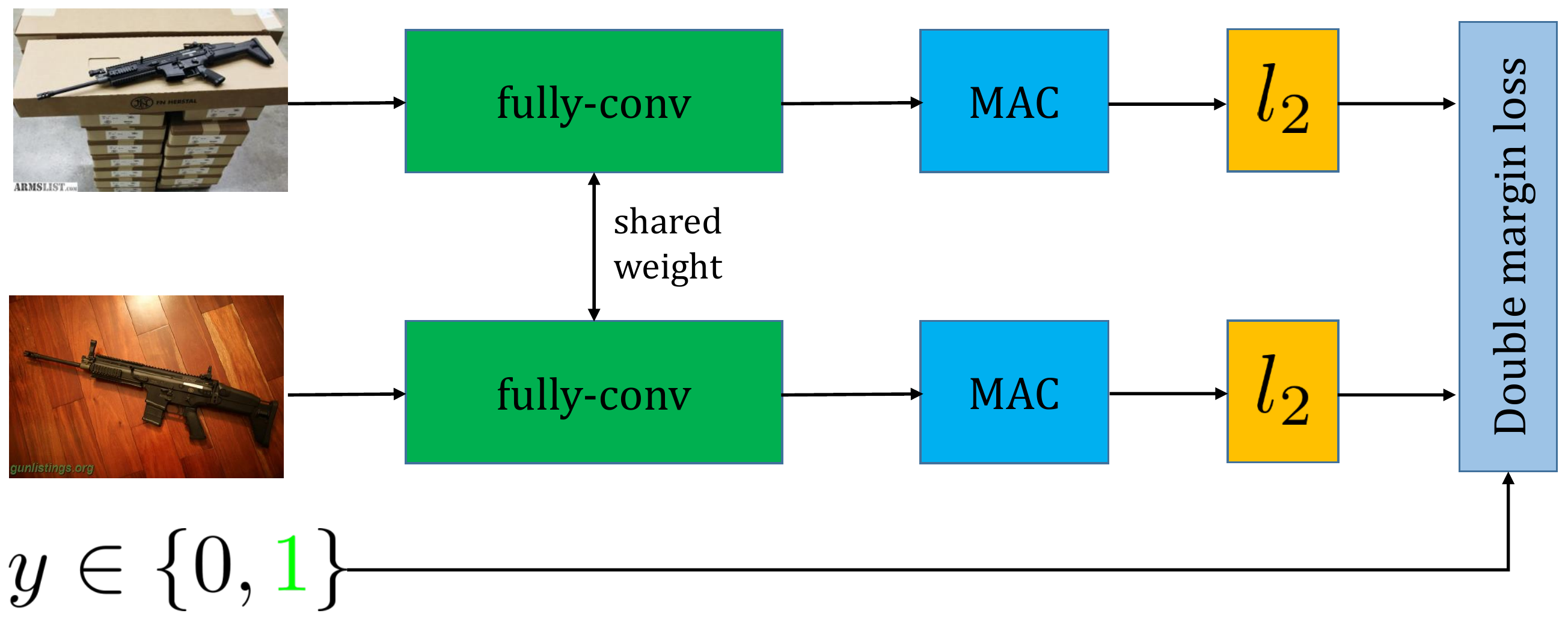}
  \caption{Architecture of our proposed Siamese network with the double margin contrastive loss. MAC~\cite{Tolias2015ParticularOR} is used to encode the image features. $l_2$ denotes $l_2$ normalization.}
  \label{fig:train_test_arch}
\end{figure}

Our model utilizes deep convolutional neural networks for generating global image feature embeddings in the Euclidean space. We use a Siamese network with two branches~\cite{Chopra2005LearningAS, Radenovic2016CNNIR} and the double margin contrastive loss to learn discriminative feature embeddings. Fig.~\ref{fig:train_test_arch} shows the training architecture of our method.

In the following section, we first describe the image feature representation method used in our work (Sec.~\ref{subsec:feature_repr}). We revisit the conventional single margin contrastive loss and introduce our improved double margin contrastive loss in Sec.~\ref{subsec:double_margin_loss}. Finally, we describe our two-stage training strategy in Sec.~\ref{subsec:staged_training}.

\subsection{Feature Representation}
\label{subsec:feature_repr}
The MAC method~\cite{Tolias2015ParticularOR} is used for the image feature representation. For an image $I$, we use the output feature maps of the last convolutional layer (after ReLU operations). We then perform global max pooling over the feature maps to obtain image features. For an image, we use the MAC-encoded feature after $l_2$-normalization.

The benefit of this feature representation is that it is fully-convolutional so that the network can accept images with different sizes and aspect ratios. We have observed an absolute 2\% performance gain on the test set by simply preserving the image aspect ratios rather than using fixed-size images during training process.

\subsection{Double Margin Contrastive Loss}\label{subsec:double_margin_loss}

First, we briefly review the conventional single margin contrastive loss. Given an image pair $(I_p, I_q)$ and their similarity label $y$.
$y=1$ if the two images are similar and $y=0$ otherwise. We denote the feature vector of image $I$ through our model as $f(I)$.
The single margin contrastive loss can be formulated as
\begin{equation}\label{eq:single_margin_loss}
L(I_p, I_q) = \frac{1}{2}\left[yd^2 + (1-y)\max(\alpha - d, 0)^2\right]\, ,
\end{equation}
where $d$ is the Euclidean distance between the $l_2$-normalized features ($d= \lVert f(I_p)-f(I_q)\rVert_2$) and $\alpha$
is the margin enforced for negative image pairs. The contrastive loss tries to pull similar images together and push dis-similar images farther than a margin $\alpha$ in the embedding space.

An apparent weakness of the single margin loss is that the network will be biased toward positive image pairs during training, as there will always be loss incurred for positive pairs, unless their feature distances are zero. Another issue is that it is hard to set a reasonable margin $\alpha$ for the negative image pairs. Both the two issues make it hard to balance the loss contributed by positive and negative image pairs.

To alleviate the issue of unbalanced loss, we propose to use a double margin contrastive loss to optimize our model. The loss being minimized is formulated as
\begin{multline}\label{eq:double_margin_loss}
L(I_p, I_q) = \frac{1}{2}[y\max(d - {\alpha}_1, 0)^2\\ + (1-y)\max({\alpha}_2 - d, 0)^2]\, ,
\end{multline}
where ${\alpha}_1$ and $\alpha_2$ are the margins for positive and negative image pairs, respectively. As can be seen from Eqn.~\ref{eq:single_margin_loss} and \ref{eq:double_margin_loss}, the single and double margin loss differ in how the loss for positive image pairs is calculated: for the former, positive image pairs always contribute to the overall
loss, while for the latter, only positive image pairs whose feature distances are larger than $\alpha_1$ contribute to the overall loss.
The double margin loss is especially suitable when we need a more balanced update to the parameters of the network to achieve further performance gains. In that case, the single margin loss is hardly effective.

\noindent\textbf{Margin Selection.} The two margin terms are among the most important hyperparameters. We choose the two margins based on the feature distance distributions of positive and negative image pairs. It is impossible to enumerate all the positive and negative pairs in the train set. In practice, we sample image pairs and choose the mean values of the two distance distributions as the starting point for selecting suitable margins. The optimal values for the two margins are then experimentally set based on the performance of the trained models on the validation set.

\subsection{Two-stage Training}\label{subsec:staged_training}
The original VGG16 model is trained on ImageNet, which is largely different from the Firearm 14k dataset. Images of the Firearm dataset are not well represented using this pretrained model. Thus, directly fine-tuning the original VGG16 model with the double margin loss will lead to sub-optimal performance. In order to alleviate the domain gap, we first fine-tune the VGG16 model for classification on the Firearm dataset. After fine-tuning the model for classification, the model are now adapted to the Firearm dataset and can already achieve relatively good retrieval performance. We proceed to optimize the model using the proposed double margin contrastive loss for the retrieval task, which further improves the performance of our model.

\section{Experiments}\label{sec:experiment}
\subsection{Settings and Implementation Details}

The PyTorch package~\cite{Pytorch2017} is employed for our whole experiments. Our model is based on the very deep VGG16 network~\cite{Simonyan2014VeryDC}. We discard all the fully-connected layers of the network to make it fully-convolutional.

For fine-tuning with classification, all 127 classes from the Firearm 14k train and validation set are utilized. We split the images into two parts for train (70\%) and validation (30\%), respectively. A weighted cross entropy loss (the weight is inversely proportional to the number of images in a class) is employed due to the highly unbalanced data. Model with the best validation accuracy is used for further fine-tuning with the double margin loss.

For fine-tuning with the double margin loss, we randomly generate an equal number (180) of positive and negative image pairs for each class in the training set. To increase the image variability and improve the generalization ability of our model, we re-generate the training image pairs every 5 epochs. During training process, we additionally augment the data by randomly resizing the larger side of an image to the range $[256, 384]$ and preserving its aspect ratio. Since the input images to the network have different sizes and aspect ratios and cannot be batch-processed, we feed one image pair to the model at a time and accumulate the loss. The parameters of the network are updated for every 64 image pairs. We do model selection on the validation set using mAP scores. SGD optimizer with momentum is utilized to optimize the model. We use an initial learning rate of 0.001 (decayed by 10 every 10 epochs), momentum 0.9 and weight decay 0.0001. The model is trained for at most 30 epochs.

\begin{table}[t]
  \centering
  \small
  \caption{Performance comparison between different components of our model and other methods.}
  \label{tab:compare_different_component}
  \begin{tabular}{@{}lllllll@{}}
    \toprule
    \multirow{2}*{} & \multirow{2}*{Method} & \multirow{2}*{mAP(\%)} & \multicolumn{4}{c}{Rank-k accuracy(\%)} \\

    \cmidrule(lr){4-7}
    & & & k=1 & k=2 & k=4 & k=8 \\
    \midrule
    (a) & VGG  & 31.1 & 83.75 & 91.25 & 95.0 & 96.25 \\
    \midrule
    (b1) & retr-s & 35.1 & 80.0 & 86.25 &  92.5&  96.25\\
    (b2) & retr-d & 47.1 & 87.5 & 88.75 & 95.0 & 97.5 \\
    \midrule
    (c1) & cls & 65.4 & 92.5 & 96.25 & 97.5 & 98.75 \\
    (c2) & cls + retr-s & -- & -- & -- & -- & -- \\
    (c3) & cls + retr-d &  68.4 & 95.0 & 98.75 &98.75 & 100.0 \\
    \bottomrule
  \end{tabular}
\end{table}

\subsection{Single VS Double Margin Contrastive Loss}
\label{subsec:single_vs_double_exp}

In this part, we evaluate the effectiveness of the proposed double margin loss (denoted as \textbf{retr-d}) over the single margin loss (denoted as \textbf{retr-s}). We report their performances in Table~\ref{tab:compare_different_component}, row (b1) and (b2). The performance of the baseline VGG16 model with MAC feature encoding (denoted as \textbf{VGG}) is also reported in row (a). The results suggest that: (1) Both the single and double margin loss can improve mAP over the baseline VGG model. (2) The rank-k accuracy metric is more challenging: the double margin loss improves the retrieval accuracy over the baseline model while the single margin loss fails on this metric. In particular, the double margin method achieves 47.1 mAP (\%) while the single margin method can only get 35.1 mAP (\%), a significant relative performance improvement (34.2\%) is gained.

\subsection{The Benefit of Two-stage Training}
\label{subsec:staged_train_exp}

We investigate the effectiveness of the two-stage training strategy in this part. We first fine-tune the VGG16 model for classification on Firearm 14k. Then we further fine-tune the model with the double margin loss. Results are reported in Table~\ref{tab:compare_different_component}, row (c1) through (c3). The fine-tuned classification model (denoted as \textbf{cls}) alone has achieved a 110.2\% performance gain over the baseline VGG16 model on the mAP metric. When we further fine-tune the classification model with the single margin loss (denoted as \textbf{cls + retr-s}), we observe that the validation performance drops consistently during training process. So the result of \textbf{cls + retr-s} on the test set is omitted. It suggests that single margin loss is not effective in improving the performance of the model further due to its unbalanced loss. In contrast, when we fine-tune the classification model with a double margin loss (denoted as \textbf{cls + retr-d}), the performance of the classification model continues to improve both in terms of mAP (4.6\% improvement) and rank-k accuracy. The results suggest that using the double margin loss is vital for the continued improvement on retrieval task.

\begin{figure}[t]
  \centering
  \includegraphics[width=\linewidth]{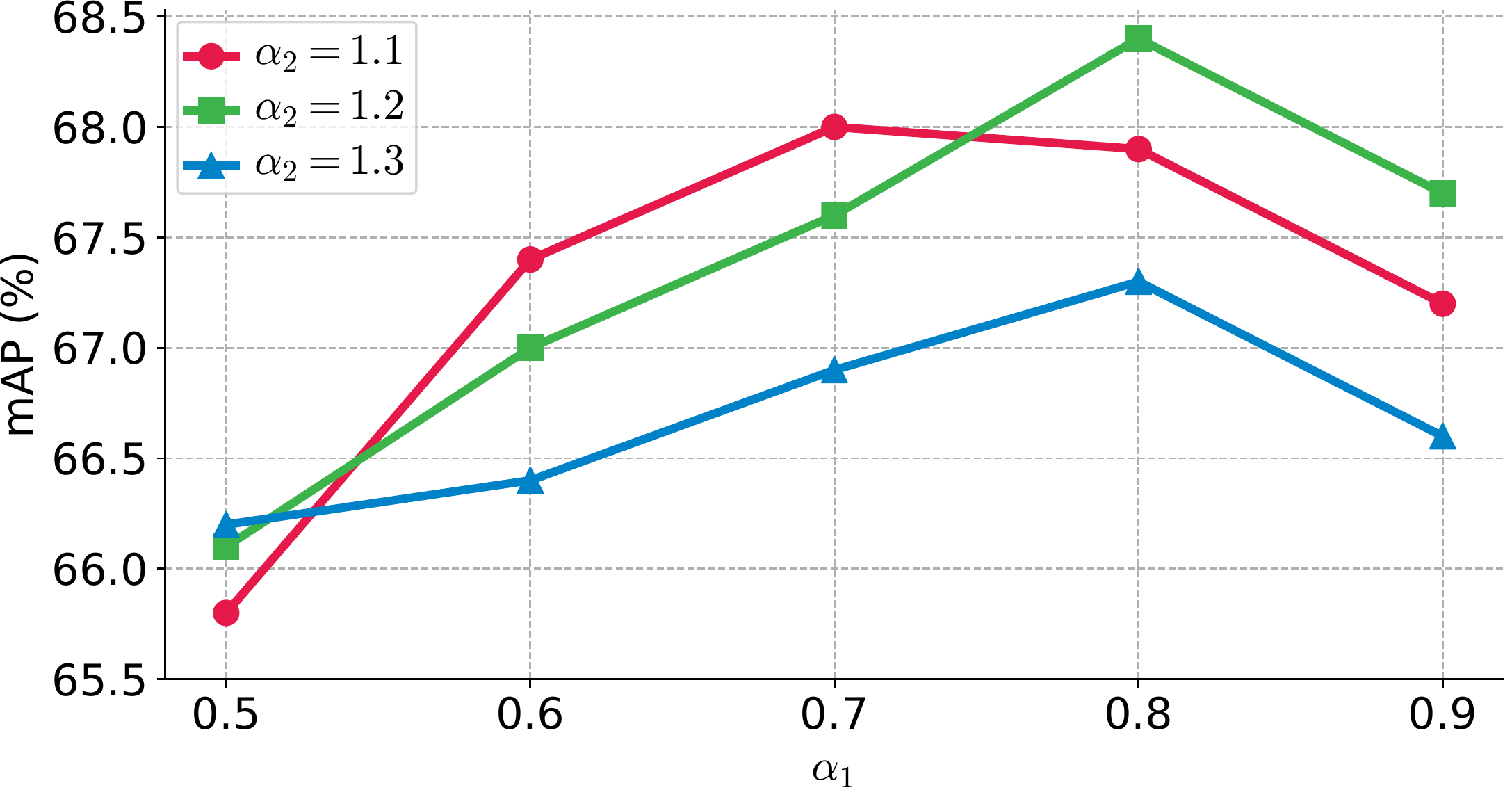}
  \caption{Impact of the two margin terms on the performance of our model.}
  \label{fig:impact_of_margins}
\end{figure}

\subsection{Impact of the Margins}\label{subsec:impact_of_margin}
\begin{figure}[t]
\centering
\includegraphics[width=\linewidth]{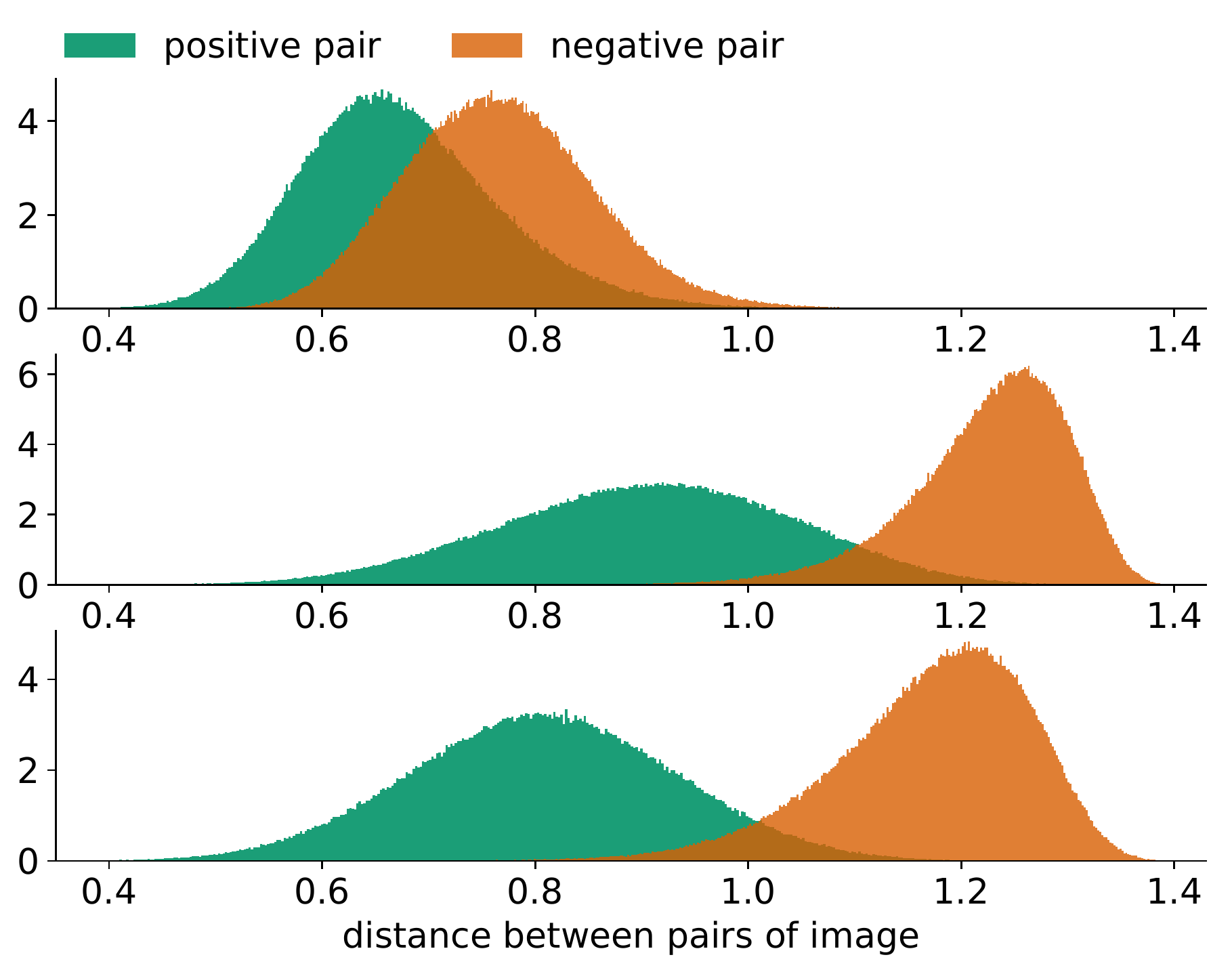}
\caption{The image pair feature distance distributions under different models. \textbf{Top}: The original VGG16 model. \textbf{Middle}: Model fine-tuned for classification on Firearm 14k. \textbf{Bottom}: Model further fine-tuned with the double margin contrastive loss based on the classification model.}
\label{fig:dist_distribution_change}
\end{figure}

In this section, we discuss the impact of the two margin terms on the performance of the model. The base model used in this experiment is the classification model in Table~\ref{tab:compare_different_component} (\textbf{cls}). We sample around 0.4M positive and negative image pairs from the dataset, calculate their features and visualize the feature distance probability distributions (Fig.~\ref{fig:dist_distribution_change}, \textbf{middle} plot). The two distributions are approximately \emph{normal distributions}. We find that the distribution mean (0.9 and 1.2 for positive and negative pairs, in this case) is a good start for choosing the margins. We then fine-tune the classification model using the double margin loss (with different margin values). Fig.~\ref{fig:impact_of_margins} shows the impact of the two margins on retrieval performance of the model. The figure suggests that the difference between the two margins should not be too big nor too small, which will all lead to under-performed models. We set $\alpha_1=0.8$ and $\alpha_2=1.2$ to optimize our final model.

\subsection{Visualizations and Qualitative Results}
\begin{figure}[t]
  \centering
  \includegraphics[width=\linewidth]{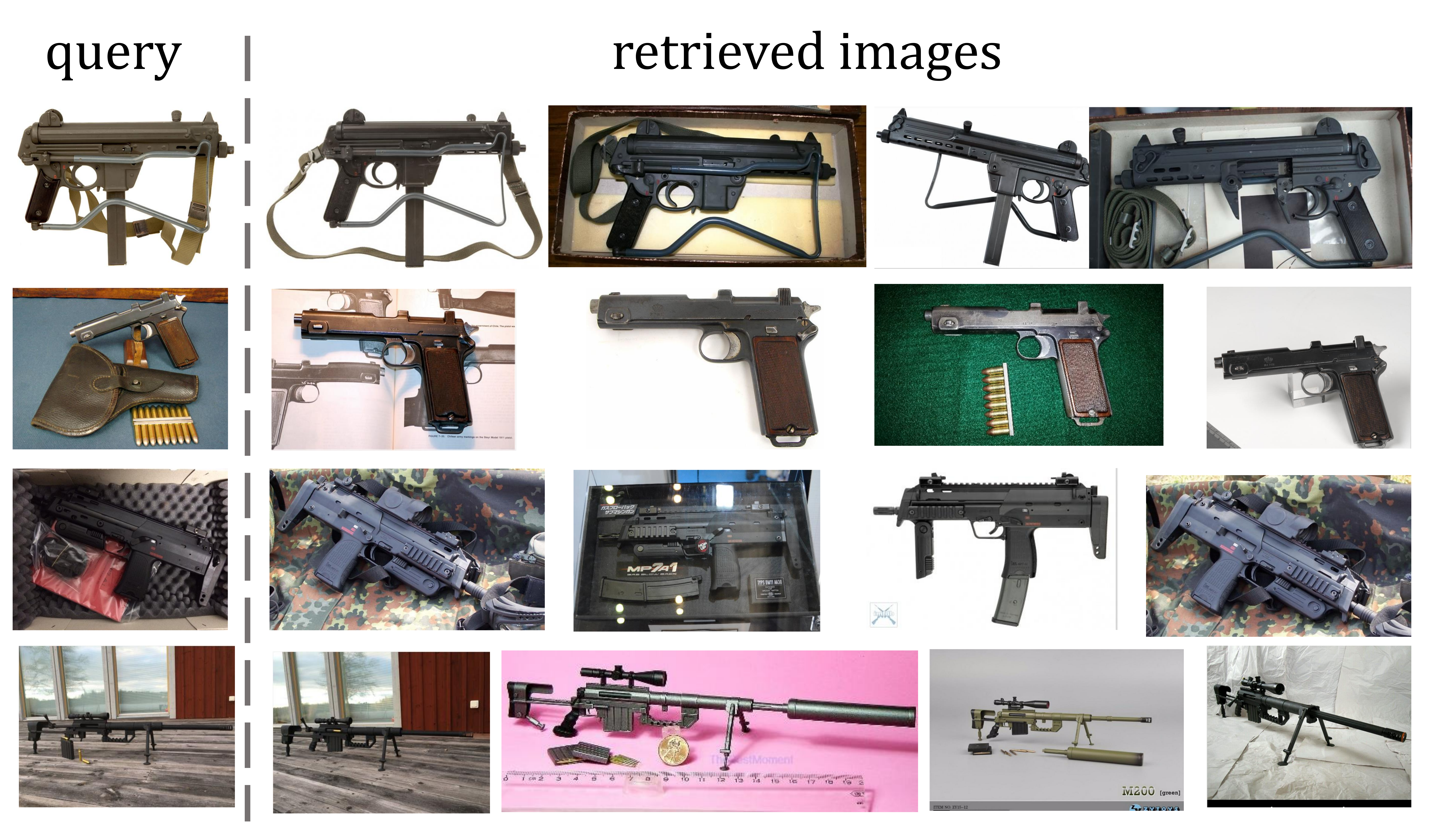}
  \caption{Top 4 retrived images for some queries on Firearm 14k.}
  \label{fig:example_query_result}
\end{figure}

We provide visualizations that help us better understand why the proposed approach is effective. Three models are compared: the original VGG16 model, model fine-tuned for classification and model initialized from the classification model and further fine-tuned with the double margin loss. For each model, we randomly sample nearly 0.5M positive and negative image pairs from the train set, respectively. The feature distance distributions of image pairs are shown in Fig.~\ref{fig:dist_distribution_change}. For the original VGG model, the positive and negative pair distance distributions largely overlap with each other (\textbf{Top} plot), suggesting that the features are not discriminative enough (mAP score 31.1\%). After fine-tuning for classification, overlap between the two distributions is reduced significantly (\textbf{Middle} plot). As a result, the mAP also improves considerably (to 65.4\%). When we further fine-tune the classification model using the double margin loss, we observe a pronounced distribution shape change for positive pairs: the shape becomes taller and thinner (\textbf{Bottom} plot) compared to its previous shape (\textbf{Middle} plot). The change in distribution shape brings about further improvement on performance (68.4\%). In Fig.~\ref{fig:example_query_result}, we also show some qualitative retrieval results of our model.

\subsection{Comparison with State of the Art}\label{subsec:compare_soa}
We compare the results of our model with several state-of-the-art methods that use either off-the-shelf or fine-tuned models. For a fair comparison, all these methods are implemented by us using the same VGG16 model. For methods using off-the-shelf models~\cite{Babenko2014NeuralCF,Tolias2015ParticularOR,Babenko_2015_ICCV}, we implement them according to the settings in the original papers. For the fine-tuned methods~\cite{Gordo2016DeepIR,Radenovic2016CNNIR}, we use the same settings as our model where possible and train these models on Firearm 14k. As is standard practice~\cite{Babenko_2015_ICCV,Radenovic2016CNNIR}, features from all these methods are $l_2$-normalized, PCA-transformed and $l_2$-normalized again.

We report the mAPs of different methods in Table~\ref{tab:compare_with_soa}. The results suggest that off-the-shelf methods (Table~\ref{tab:compare_with_soa}, top) are largely ineffective as their mAP scores are relatively low. This is due to the fact that the Firearm 14k dataset is different from ImageNet where the model is trained. The pretrained model cannot generate discriminative features for images in this dataset. Fine-tuned models using single margin contrastive loss~\cite{Radenovic2016CNNIR} also performs poorly. The TripletNet~\cite{Gordo2016DeepIR} performs better compared to the single margin method. Finally, our proposed model outperforms all these methods under different feature dimensionality. It is also worth noting that the PCA-transformed features work surprisingly well even if the feature dimension is compressed to only 64.

\begin{table}[t]
  \centering
  \begin{threeparttable}
    \caption{Comparison with state-of-the-art methods.}
  \label{tab:compare_with_soa}
  \begin{tabular}{@{}lllll@{}}
    \toprule
   \multirow{2}*{Method} & \multicolumn{4}{c}{Feature dimension}\\
    \cmidrule(lr){2-5}
     & D=512 & D=256 & D=128 & D=64 \\
    \midrule
    Neural codes~\cite{Babenko2014NeuralCF} & 15.9 & 15.8 & 15.7 & 14.6 \\
    SPoC~\cite{Babenko_2015_ICCV} & 23.3 & 23.4 & 23.1 & 22.4 \\
    MAC~\cite{Tolias2015ParticularOR} & 36.1 & 36.4 & 36.5 & 35.4 \\
    \midrule
    Siamese-MAC~\cite{Radenovic2016CNNIR} & 35.6 & 36.2 & 37.2 & 37.3 \\
    TripletNet\textsuperscript{$\dagger$}~\cite{Gordo2016DeepIR} & 67.98 & 68.57 & 69.60 & 69.97 \\
    \midrule
    Ours (retr-d) & 45.79 & 46.17 & 46.61 & 46.75  \\
    Ours (cls) & 66.57 & 67.2 & 68.53 & 68.67 \\
    Ours best (cls + retr-d) & \textbf{68.63} & \textbf{69.15} & \textbf{70.14} & \textbf{70.07} \\
    \bottomrule
  \end{tabular}
  \begin{tablenotes}
      \footnotesize
      \item \textsuperscript{$\dagger$} The TripletNet is also initialized from the classification model fine-tuned on Firearm 14k.
  \end{tablenotes}
  \end{threeparttable}
\end{table}

\section{Conclusion}
In this work, we present Firearm 14k, a large dataset of firearm images for research on fine-grained firearm recognition and retrieval. We propose to use the double margin contrastive loss to alleviate the problem of unbalanced loss between positive and negative image pairs when the conventional single margin loss is employed. We further develop a two-stage training strategy to deal with the large domain gap between the Firearm 14k dataset and ImageNet. We conduct extensive experiments and give detailed analysis and visualizations showing that our model achieves superior retrieval performance under different feature dimensionality compared to existing state-of-the-art methods.

\section*{Acknowledgment}
This work is supported by National Natural Science Foundation of China under Grant No. U1536120, U1636201, U1736119, 61502496 and 61772529, the National Key Research and Development Program of China under Grant No. 2016YFB1001003.


\bibliographystyle{IEEEtran}
\bibliography{references.bib}

\begin{thebibliography}{10}
\providecommand{\url}[1]{#1}
\csname url@samestyle\endcsname
\providecommand{\newblock}{\relax}
\providecommand{\bibinfo}[2]{#2}
\providecommand{\BIBentrySTDinterwordspacing}{\spaceskip=0pt\relax}
\providecommand{\BIBentryALTinterwordstretchfactor}{4}
\providecommand{\BIBentryALTinterwordspacing}{\spaceskip=\fontdimen2\font plus
\BIBentryALTinterwordstretchfactor\fontdimen3\font minus
  \fontdimen4\font\relax}
\providecommand{\BIBforeignlanguage}[2]{{%
\expandafter\ifx\csname l@#1\endcsname\relax
\typeout{** WARNING: IEEEtran.bst: No hyphenation pattern has been}%
\typeout{** loaded for the language `#1'. Using the pattern for}%
\typeout{** the default language instead.}%
\else
\language=\csname l@#1\endcsname
\fi
#2}}
\providecommand{\BIBdecl}{\relax}
\BIBdecl

\bibitem{Forbes2016FacebookGun}
\url{https://www.forbes.com/sites/mattdrange/2016/03/31/facebook-guns-beet_farmer-image-recognition}.

\bibitem{Krizhevsky2012ImageNetCW}
A.~Krizhevsky, I.~Sutskever, and G.~E. Hinton, ``Imagenet classification with
  deep convolutional neural networks,'' in \emph{NIPS}, 2012, pp. 1097--1105.

\bibitem{Russakovsky2015ImageNetLS}
O.~Russakovsky, J.~Deng, H.~Su, J.~Krause, S.~Satheesh, S.~Ma, Z.~Huang,
  A.~Karpathy, A.~Khosla, M.~S. Bernstein, A.~C. Berg, and L.~Fei-Fei,
  ``Imagenet large scale visual recognition challenge,'' \emph{IJCV}, vol. 115,
  pp. 211--252, 2015.

\bibitem{Ren2015FasterRT}
S.~Ren, K.~He, R.~B. Girshick, and J.~Sun, ``Faster r-cnn: Towards real-time
  object detection with region proposal networks,'' \emph{PAMI}, vol.~39, pp.
  1137--1149, 2015.

\bibitem{Shelhamer2015FullyCN}
E.~Shelhamer, J.~Long, and T.~Darrell, ``Fully convolutional networks for
  semantic segmentation,'' \emph{PAMI}, vol.~39, pp. 640--651, 2015.

\bibitem{Radenovic2016CNNIR}
F.~Radenovic, G.~Tolias, and O.~Chum, ``Cnn image retrieval learns from bow:
  Unsupervised fine-tuning with hard examples,'' in \emph{ECCV}, 2016, pp.
  3--20.

\bibitem{Chopra2005LearningAS}
S.~Chopra, R.~Hadsell, and Y.~LeCun, ``Learning a similarity metric
  discriminatively, with application to face verification,'' in \emph{CVPR},
  2005, pp. 539--546.

\bibitem{Gordo2016DeepIR}
A.~Gordo, J.~Almaz{\'a}n, J.~Revaud, and D.~Larlus, ``Deep image retrieval:
  Learning global representations for image search,'' in \emph{ECCV}, 2016, pp.
  241--257.

\bibitem{Wang2014LearningFI}
J.~Wang, Y.~S. Song, T.~Leung, C.~Rosenberg, J.~Wang, J.~Philbin, B.~Chen, and
  Y.~Wu, ``Learning fine-grained image similarity with deep ranking,'' in
  \emph{CVPR}, 2014, pp. 1386--1393.

\bibitem{Schroff2015FaceNetAU}
F.~Schroff, D.~Kalenichenko, and J.~Philbin, ``Facenet: A unified embedding for
  face recognition and clustering,'' in \emph{CVPR}, 2015, pp. 815--823.

\bibitem{Bell2015LearningVS}
S.~Bell and K.~Bala, ``Learning visual similarity for product design with
  convolutional neural networks,'' \emph{TOG}, vol.~34, pp. 98:1--98:10, 2015.

\bibitem{Philbin2007ObjectRW}
J.~Philbin, O.~Chum, M.~Isard, J.~Sivic, and A.~Zisserman, ``Object retrieval
  with large vocabularies and fast spatial matching,'' in \emph{CVPR}, 2007,
  pp. 1--8.

\bibitem{Philbin2008LostIQ}
------, ``Lost in quantization: Improving particular object retrieval in large
  scale image databases,'' in \emph{CVPR}, 2008, pp. 1--8.

\bibitem{Tolias2015ParticularOR}
G.~Tolias, R.~Sicre, and H.~J{\'e}gou, ``Particular object retrieval with
  integral max-pooling of cnn activations,'' \emph{CoRR}, vol. abs/1511.05879,
  2015.

\bibitem{Babenko_2015_ICCV}
A.~Babenko and V.~Lempitsky, ``Aggregating local deep features for image
  retrieval,'' in \emph{ICCV}, 2015, pp. 1269--1277.

\bibitem{Babenko2014NeuralCF}
A.~Babenko, A.~Slesarev, A.~Chigorin, and V.~Lempitsky, ``Neural codes for
  image retrieval,'' in \emph{ECCV}, 2014, pp. 584--599.

\bibitem{Simonyan2014VeryDC}
K.~Simonyan and A.~Zisserman, ``Very deep convolutional networks for
  large-scale image recognition,'' \emph{CoRR}, vol. abs/1409.1556, 2014.

\bibitem{Wen2005PistolIR}
C.-Y. Wen and J.-Y. Yao, ``Pistol image retrieval by shape representation.''
  \emph{Forensic science international}, vol. 155 1, pp. 35--50, 2005.

\bibitem{Razavian2014CNNFO}
A.~S. Razavian, H.~Azizpour, J.~Sullivan, and S.~Carlsson, ``Cnn features
  off-the-shelf: An astounding baseline for recognition,'' in \emph{CVPRW},
  2014, pp. 512--519.

\bibitem{kaiming14ECCV}
K.~He, X.~Zhang, S.~Ren, and J.~Sun, ``Spatial pyramid pooling in deep
  convolutional networks for visual recognition,'' in \emph{ECCV}, 2014, pp.
  346--361.

\bibitem{Hao2017MFCAM}
J.~Hao, W.~Wang, J.~Dong, and T.~Tan, ``{MFC}: A multi-scale fully
  convolutional approach for visual instance retrieval,'' in \emph{ICME
  Workshops}, 2017.

\bibitem{Cao2016QuartetnetLF}
J.~Cao, Z.~Huang, P.~Wang, C.~Li, X.~Sun, and H.~T. Shen, ``Quartet-net
  learning for visual instance retrieval,'' in \emph{ACM MM}, 2016, pp.
  456--460.

\bibitem{Pytorch2017}
\url{http://pytorch.org/}.

\end{thebibliography}

\end{document}